\documentclass{article}

\PassOptionsToPackage{numbers}{natbib}
\usepackage[preprint]{neurips_2024}

\usepackage[utf8]{inputenc} 
\usepackage[T1]{fontenc}    
\usepackage{hyperref}       
\usepackage{url}            
\usepackage{booktabs}       
\usepackage{amsfonts}       
\usepackage{nicefrac}       
\usepackage{microtype}      
\usepackage{xcolor}         
\usepackage[utf8]{inputenc} 
\usepackage[T1]{fontenc}    
\usepackage{hyperref}       
\usepackage{url}            
\usepackage{booktabs}       
\usepackage{amsfonts}       
\usepackage{nicefrac}       
\usepackage{microtype}      
\usepackage{graphicx}
\usepackage{xcolor}         
\usepackage{amsmath}
\usepackage{amssymb} 
\usepackage{algorithm}
\usepackage{algorithmic}
\usepackage{subcaption}
\usepackage{bm}
\usepackage{acro}
\DeclareMathOperator*{\argmax}{arg\,max}

\title{Improving Generalization of Robot Locomotion \\
Policies via Sharpness-Aware Reinforcement Learning}

\author{
    S.~Bochem$^{1}$\textsuperscript{*}, 
    E.~Gonzalez-Sanchez$^{1,2}$\textsuperscript{*,\dag}, 
    Y.~Bicker$^{1,3}$\textsuperscript{*}, 
    G.~Fadini$^{4}$ \\
    $^1$ETH Zürich, Zürich, Switzerland \\
    $^2$inspire AG, Zürich, Switzerland \\
    $^3$University of Zurich, Zürich, Switzerland \\
        $^4$ETH Zürich Computational Robotics Lab (CRL), Zürich, Switzerland \\
    \textsuperscript{*}\,Equal contribution. \\
    \textsuperscript{\dag}\,Corresponding author: \href{mailto:geduardo@ethz.ch}{\texttt{geduardo@ethz.ch}}
}

\begin{document}

\maketitle

\begin{abstract}
    Reinforcement learning often requires extensive training data. Simulation-to-real transfer offers a promising approach to address this challenge in robotics. While differentiable simulators offer improved sample efficiency through exact gradients, they can be unstable in contact-rich environments and may lead to poor generalization. This paper introduces a novel approach integrating sharpness-aware optimization into gradient-based reinforcement learning algorithms.
    Our simulation results demonstrate that our method, tested on contact-rich environments, significantly enhances policy robustness to environmental variations and action perturbations while maintaining the sample efficiency of first-order methods. Specifically, our approach improves action noise tolerance compared to standard first-order methods and achieves generalization comparable to zeroth-order methods.
    This improvement stems from finding flatter minima in the loss landscape, associated with better generalization. Our work offers a promising solution to balance efficient learning and robust sim-to-real transfer in robotics, potentially bridging the gap between simulation and real-world performance.
\end{abstract}

\section{Introduction}
Reinforcement Learning (RL) has been successfully employed to learn robust control policies for robotic environments from data.
A major downside of RL is the large amount of training experience it needs to approximate the policy gradient, which may become unfeasible online.
In response to this challenge, transfer learning in robotics allows for policy development in simulation before deployment on real robots \cite{Hutter, openai2019learningdexterousinhandmanipulation}, bypassing the challenges of direct real-world learning \cite{tan2018simtoreallearningagilelocomotion}. However, the gap between simulations and the real world remains a significant challenge \cite{tobin_domain_2017}.
Differentiable simulators have emerged as powerful tools for sample-efficient policy optimization, enabling the use of first-order policy gradient (FoPG) in policy training \cite{PODS}.
These methods leverage analytic gradients of a policy's value function, leading to faster convergence and improved sample efficiency compared to zeroth-order methods. However, the effectiveness of FoPG methods relies heavily on the quality and accuracy of simulator gradients.
In real-world robotics applications, particularly those involving complex contact interactions, the dynamics are often non-differentiable \cite{lidec2024contactmodelsroboticscomparative, 8594284} producing rugged landscapes with sharp local minima \cite{antonova2022rethinkingoptimizationdifferentiablesimulation}.
Approximations used to make simulators differentiable can introduce bias and high variance in the computed gradients \cite{suh2022differentiable}.
The trade-off between using gradient-based information and zeroth-order methods in reinforcement learning remains an open question in the field of policy optimization for robotics.
Gradient-based methods typically offer more efficient parameter updates but may struggle with non-smooth or discrete action spaces,
whereas zero-order methods seem to handle these scenarios more easily but may require more samples. Additionally, the generalization capabilities of first-order policy optimization methods, which have become increasingly popular, still need to be thoroughly tested across a wide range of environments and tasks.
As the field progresses, understanding the relative strengths and limitations of these approaches in different contexts remains an important area of research that we aim to address.

The rest of the paper is structured as follows. The following section \ref{sec:related_work} discusses recent work on differentiable simulation and the generalizability of trained policies. Section \ref{sec:background} briefly introduces the necessary background of this paper. In the methods section \ref{sec:algorithm} we formally introduce our proposed method to improve the robustness of policies trained with differentiable simulation. In section \ref{sec:results}, we showcase that policies trained with our proposed algorithm are able to improve generalizability. Finally, we give an outlook on the limitations of our work and future research directions.

\section{Related Work} \label{sec:related_work}

Recent works such as Brax \cite{freeman_brax_2021}, DiffPD \cite{du_diffpd_2021}, Dojo \cite{howell2023dojo}, and ADD \cite{geilinger2020addanalyticallydifferentiabledynamics} provide differentiable simulators to enhance sample-efficiency of policy optimization in robotics.
Building upon this foundation, algorithms like Policy Optimization with Differentiable Simulation (PODS) \cite{PODS} and Short Horizon Actor Critic (SHAC) \cite{xu2022accelerated} have leveraged the analytical gradients provided by differentiable simulators to significantly improve sample efficiency compared to model-free methods like PPO.
SHAC, in particular, has made strides in addressing the challenges of contact-rich dynamics through techniques like truncated learning windows and critic function smoothing.
More recently, Adaptive Horizon Actor-Critic (AHAC) \cite{georgiev2024adaptivehorizonactorcriticpolicy} has further refined this approach by dynamically adjusting the optimization horizon based on contact information.

For successful transfer to unknown environments, learned policies must be robust to the discrepancies between simulated and real-world environments.
Prior work has approached this challenge from various angles, such as domain randomization \cite{caluwaerts2023barkourbenchmarkinganimallevelagility} and domain adaptation \cite{9327467}. 

Previous work in deep learning introduces the concept of flatter minima in the optimization landscape and shows that they lead to more generalizable models \cite{hochreiter1997flat}.
Furthermore, empirical studies in deep learning have shown that models converging to these minima tend to exhibit better out-of-distribution performance \cite{keskar2017largebatchtrainingdeeplearning}, a property highly desirable for sim-to-real transfer in robotics.
A technique to actively search for these flatter minima is Sharpness-Aware Minimization (SAM) \cite{foret_sharpness-aware_2021} and its adaptive counterpart, Adaptive Sharpness Aware Minimization (ASAM) \cite{kwon2021asam}. As these optimizers require two backward passes in each optimization step, a later work, Sharpness-Aware Training for Free (SAF) proposes a sharpness measure based on the KL-divergence between the outputs of DNNs with the current weights and past weights, overcoming the high computational cost of ASAM \cite{SAF}. Efficient Sharpeness-Aware Minimization (ESAM) improves the computational overhead of ASAM from 100 $\%$ to 40 $\%$ \cite{du2022efficient}. 
To the best of our knowledge, none of the existing FoPG algorithms have incorporated this insight.

\section{Contributions}
In this work, we demonstrate that while the first-order policy method SHAC \cite{xu2022accelerated} is more sample-efficient and achieves better rewards than the zeroth-order method Proximal Policy Optimization (PPO) \cite{schulman2017proximal}, it struggles with generalization, particularly in noisy and out-of-distribution environments.
To address this robustness issue, we introduce a novel approach SHAC-ASAM that incorporates sharpness-aware optimizers \cite{kwon2021asam} into the training process of first-order methods.
Our experimental results demonstrate that SHAC-ASAM significantly enhances the robustness of policies compared to vanilla SHAC in both the Ant and Humanoid environments, effectively bridging the gap between first-order efficiency and zeroth-order robustness.
Our approach balances sample efficiency with generalization, crucial for developing policies that can effectively navigate the sim-to-real gap in robotics.
By combining the rapid learning of first-order methods with an enhanced ability to generalize, we aim to create policies that are both efficient to train and robust in real-world applications.

\section{Preliminaries} \label{sec:background}

\subsection{Differentiable Simulation gradients}
A differentiable simulator defines a differentiable function $\mathbf{s}_{t+1} = \mathcal{F}(\mathbf{s}_t, \mathbf{a}_t)$, mapping the current state $\mathbf{s}_t$ and action $\mathbf{a}_t$ to the next state $\mathbf{s}_{t+1}$. During the forward pass, the simulator generates a trajectory by applying the policy $\pi_\theta$ and simulating the environment dynamics using $\mathcal{F}$. During the backward pass, it computes the gradients of the policy loss $\mathcal{L}_\theta$ with respect to $\mathbf{s}_t$ and $\mathbf{a}_t$:

\begin{equation}
\frac{\partial \mathcal{L}_\theta}{\partial \mathbf{s}_t} = \frac{\partial \mathcal{L}_\theta}{\partial \mathbf{s}_{t+1}} \frac{\partial \mathcal{F}}{\partial \mathbf{s}_t}, \quad \frac{\partial \mathcal{L}_\theta}{\partial \mathbf{a}_t} = \frac{\partial \mathcal{L}_\theta}{\partial \mathbf{s}_{t+1}} \frac{\partial \mathcal{F}}{\partial \mathbf{a}_t}.
\end{equation}

A differentiable simulator enables efficient policy gradient computation by backpropagating the policy loss through the simulator's computation graph, resulting in exact, low-variance gradients for faster convergence and more stable optimization.
In this work, we aim to improve the robustness of policies learned by algorithms that leverage differentiable simulations.

\subsection{Short Horizon Actor-Critic}
SHAC leverages analytical gradients from a differentiable simulator to address the challenges of contact-rich dynamics, long horizons, and sample efficiency in reinforcement learning \cite{xu2022accelerated}. The algorithm splits the task horizon into sub-windows of smaller horizons and samples $N$ short-horizon trajectories of length $h \ll H$ in parallel from the simulator, where $H$ is the full task horizon. The policy is updated with:

\begin{equation}
\mathcal{L}_{\theta}=-\frac{1}{Nh}\sum_{i=1}^{N}\bigg{[}\Big{(} \sum_{t=t_{0}}^{t_{0}+h-1}\gamma^{t-t_{0}}\mathcal{R}(\mathbf{s}_{t}^{i}, \mathbf{a}_{t}^{i})\Big)+\gamma^{h}V_{\phi}(\mathbf{s}_{t_{0}+h}^{i})\bigg]
\label{eq:loss_shac_detailed}
\end{equation}

where $\mathbf{s}_{t}^{i}$ and $\mathbf{a}_{t}^{i}$ are the state and action at step $t$ of the $i$-th trajectory, $\gamma$ is the discount factor, $V_{\phi}$ is the critic function with parameters $\phi$, $\mathcal{R}(\mathbf{s}_{t}^{i}, \mathbf{a}_{t}^{i})$ is the reward function, $t_0$ is the initial time step of the short horizon, and $h$ is the length of the short horizon. The short horizon reduces the effect of exploding/vanishing gradients and helps deal with severe discontinuities, leading to a smoother loss landscape. The value function is trained using the following loss:

\begin{equation}
\mathcal{L}_{\phi}=\underset{\mathbf{s}\in\{\tau_{i}\}}{\mathbb{E}}\bigg{[} \|V_{\phi}(\mathbf{s})-\tilde{V}(\mathbf{s})\|^{2}\bigg{]}
\label{eq:loss_shac}
\end{equation}

where $\tilde{V}(\mathbf{s})$ is the estimated target value of the true value function for state $\mathbf{s}$, computed from the sampled short-horizon trajectories using a suitable algorithm like TD($\lambda$) learning. Here, $\tau_i$ represents an individual trajectory, which is a sequence of states, actions, and rewards experienced by the agent during a single episode or rollout of the environment.

\subsection{Adaptive Sharpness-Aware Minimization}

ASAM \cite{kwon2021asam} improves model generalization by minimizing loss value and loss sharpness of the parameter space simultaneously. The objective is a minimax optimization problem:

\begin{equation}
\begin{aligned}
    \min_{\mathbf{\theta}} \quad & L_{\mathcal{S}}^{ASAM}(\mathbf{\theta}) + \lambda \|\mathbf{\theta}\|_{2}^{2} \quad, \\
    \text{where:} & \quad L_{\mathcal{S}}^{ASAM}(\mathbf{\theta}) \triangleq \max_{\|\mathbf{T_\mathbf{\theta}^{-1} \mathbf{\epsilon}}\|_{p} \leq \rho} L_{\mathcal{S}}(\mathbf{\theta} + \mathbf{\epsilon}),
\end{aligned}
\label{eq:sam}
\end{equation}

The inner maximization of ASAM uses a scale-invariant perturbation $\mathbf{\epsilon}$ within an $\ell_p$ norm ball of radius $\rho$ to maximize the training loss $L_{\mathcal{S}}(\mathbf{\theta})$, identifying the worst-case scenario. The maximization can be interpreted as finding the highest loss value corresponding to sharp peaks within the local neighborhood. The outer minimization updates $\mathbf{\theta}$ to minimize this worst-case loss, promoting flatter minima in the loss landscape, which leads to better generalization performance. The normalization operator $T_\mathbf{\theta}^{-1}$ adjusts the neighborhood's size and shape based on model parameters, ensuring invariance to parameter scaling. This maximization is approximated by a first-order Taylor expansion around $\mathbf{\theta}$, yielding a closed-form solution.

\begin{align}
\tilde{\boldsymbol{\epsilon}}_t 
&= \argmax_{\|\tilde{\boldsymbol{\epsilon}}\|_p \leq \rho} L_S(\mathbf{\theta}_t + T_{\mathbf{\theta}_t}\tilde{\boldsymbol{\epsilon}})\\  &\approx \rho \operatorname{sign}(\nabla L_S(\mathbf{\theta}_t)) \frac{|T_{\mathbf{\theta}_t} \nabla L_S(\mathbf{\theta}_t)|^{q-1}}{\|T_{\mathbf{\theta}_t} \nabla L_S(\mathbf{\theta}_t)\|_q^{q-1}}
\end{align}

where $\tilde{\boldsymbol{\epsilon}} = T_{\mathbf{\theta}}^{-1}\boldsymbol{\epsilon}$. When applying ASAM to SHAC, the perturbation term is computed from the gradient of the loss function in equation \ref{eq:loss_shac_detailed}, with respect to $\nabla_\mathbf{\theta} L_S(\mathbf{\theta}_t)$, and the two-step procedure iteratively solves the minimax optimization problem:

\begin{equation}
\left\{
\begin{aligned}
\boldsymbol{\epsilon}_t &= \rho T_{\mathbf{\theta}_t} \operatorname{sign}(\nabla L_S(\mathbf{\theta}_t)) \frac{|T_{\mathbf{\theta}_t} \nabla L_S(\mathbf{\theta}_t)|^{q-1}}{\|T_{\mathbf{\theta}_t} \nabla L_S(\mathbf{\theta}_t)\|_q^{q-1}} \\
\mathbf{\theta}_{t+1} &= \mathbf{\theta}_t - \alpha_t (\nabla L_S(\mathbf{\theta}_t + \boldsymbol{\epsilon}_t) + \lambda \mathbf{\theta}_t)
\end{aligned}
\right.
\end{equation}

for $t = 0, 1, 2, \ldots$, where $\alpha_t$ is the learning rate and $\rho$ controls the norm-ball radius and hence the perturbation strength. 
Using the Euclidean norm ($p = 2$) and normalization operator $T_{\mathbf{\theta}_t} = \text{diag}(|\mathbf{\theta}_t|)$, the perturbation term simplifies to:

\begin{equation}
\boldsymbol{\epsilon}_t = \rho \frac{T_{\mathbf{\theta}_t}^2 \nabla L_S(\mathbf{\theta}_t)}{\|T_{\mathbf{\theta}_t} \nabla L_S(\mathbf{\theta}_t)\|_2}
\end{equation}
\\

While our experiments focus on SHAC, it is important to note that the principles underlying our approach are not limited to this specific algorithm. The integration of sharpness-aware optimization techniques should, in principle, apply to other algorithms that leverage the differentiability of the underlying simulator, such as AHAC\cite{georgiev2024adaptivehorizonactorcriticpolicy} and PODS \cite{PODS}. 


\section{SHAC-ASAM Algorithm}
\label{sec:algorithm}
 Our algorithm SHAC-ASAM is detailed in Alg.~\ref{alg:SHAC-ASAM}. It combines SHAC's sample efficiency with ASAM's robustness, resulting in stable policies for contact-rich, long-horizon tasks with limited samples, potentially improving sim-to-real transfer.
In every episode of the policy optimization, the minimax optimization problem \eqref{eq:sam} is solved.
Our SHAC-ASAM approach utilizes the ASAM optimizer proposed by \cite{kwon2021asam}, which offers improved scale-invariance.
This extension leverages the unofficial PyTorch repository \cite{david_sam_2020}, which provides a straightforward way of integrating SAM and ASAM into existing PyTorch-based pipelines.
In practice, the SAM optimizer acts as a wrapper around a base optimizer
(in our case, Adam \cite{kingma2017adammethodstochasticoptimization}), computing a "sharpness-aware" gradient that simultaneously minimizes both the loss value and its sharpness.
This approach seeks parameters that lie in neighborhoods with uniformly low loss, potentially leading to better generalization.
The ASAM variant adapts the neighborhood size based on the parameter scale, further enhancing robustness.

The optimization problem in \eqref{eq:sam} is more complex than the one solved in vanilla SHAC and hence the integration of ASAM leads to additional computational overhead.
In particular, two forward-backward passes per optimization step are required, as outlined in algorithm \ref{alg:SHAC-ASAM}.
While the computational cost is increased, the potential improvements in generalization and robustness make it a valuable tool for reinforcement learning tasks, particularly those aimed at sim-to-real transfer.

\begin{algorithm}
\caption{SHAC-ASAM Policy Learning}\label{alg:SHAC-ASAM}
\begin{algorithmic}[1]
\STATE Initialize policy $\pi_\theta$, value function $V_\phi$, and target value function $V_{\phi_0} \leftarrow V_\phi$.
\FOR{learning episode $= 1, 2, \dots, M$}
\STATE Sample $N$ short-horizon trajectories of length $h$ by the parallel differentiable simulation from the final states of the previous trajectories.
\STATE Compute the SHAC policy loss $L_{\theta_t}$ defined in \eqref{eq:loss_shac_detailed} from the sampled trajectories and $V_{\phi_0}$.
\STATE Compute the analytical gradient $\nabla L(\theta_t)$.
\STATE Update the policy $\pi_{\theta_t}$ using ASAM:
\begin{enumerate}
\item Determine the normalization operator \( T_{\theta}^{-1} \) to adjust the neighborhood size based on model parameters
\item Compute the perturbation ${\epsilon}_t$ and update the model parameters temporarily:
\begin{align}
\bm{\epsilon}_t &= \rho \frac{T^2{\theta_t} \nabla L(\theta_t)}{\|T_{\theta_t} \nabla L(\theta_t)\|_2} & \tilde{\mathbf{\theta}}_t = \mathbf{\theta}_t + {\epsilon}_t
\end{align}
\item Update the model parameters using the gradient at the perturbed point:
\begin{equation}
\mathbf{\theta}_{t+1} = \mathbf{\theta}_t - \alpha_t (\nabla L(\tilde{\mathbf{\theta}}_t) + \lambda \cdot\mathbf{\theta_t})
\end{equation}
\end{enumerate}
\STATE Compute estimated values $\tilde{V}(\mathbf{s})$ for all the states in sampled trajectories with the td-$\lambda$ formulation in \cite{xu2022accelerated}.
\STATE Fit the value function $V_\phi$ using the critic loss defined in \eqref{eq:loss_shac}.
\STATE Update target value function: $V_{\phi_0} \leftarrow \alpha V_{\phi_0} + (1 - \alpha)V_\phi$.
\ENDFOR
\end{algorithmic}
\end{algorithm}

\section{Environment Perturbation}
\subsection{Noise Injection Mechanism on Actions} \label{sub:noise_actions}

To evaluate the robustness of the learned policies against controlled perturbations, we introduce an action noise injection mechanism.
It is important to note that although this action noise is not necessarily reflecting perturbation sources from the real world (e.g. friction, delays, actuator dynamics), it acts as an upper bound, allowing us a fine control of the perturbation in the policy actions.
The original actions $\mathbf{a}$ are clipped to the range $[-1, 1]$:
\begin{equation}
\mathbf{a} = \text{clip}(\mathbf{a}, -1, 1)
\end{equation}
We then generate uniform noise $\mathbf{n}$ sampled from $\mathcal{U}(-1, 1)$, and compute noisy actions $\mathbf{a}'$ as a convex combination of $\mathbf{a}$ and $\mathbf{n}$, controlled by $\lambda \in [0, 1]$:
\begin{equation}
\mathbf{a}' = (1 - \lambda) \mathbf{a} + \lambda \mathbf{n}
\end{equation}
The parameter $\lambda$ determines the noise strength, with ${\lambda = 0}$ corresponding to no noise and ${\lambda = 1}$ to complete replacement of the original action by noise. The convex combination ensures that $\mathbf{a}'$ remains within $[-1, 1]$.
This mechanism allows us to systematically assess the robustness of the learned policies under varying levels of controlled noise. This approach provides insights into policy behavior under controlled perturbations. Policies that maintain good performance even under stronger noise injection are considered more robust, indicating their better generalization in perturbed scenarios.

\subsection{Environment Parameter Variation for Sim-to-Real Transfer Assessment}\label{sub:contac_params}

As an additional mechanism to evaluate policy robustness, we systematically vary key simulation parameters, aiming to approximate sim-to-real transfer:

\begin{itemize}
    \item Contact stiffness ($k_e$): Affects the rigidity of interactions between the robot and its environment
    \item Coefficient of Friction ($\mu$): Influences the force required for surfaces to slide against each other
    \item Contact damping ($k_d$): Impacts energy dissipation during contact
\end{itemize}

This approach simulates out-of-distribution environments, assesses policy generalization to shifted physical properties, provides insights into potential real-world performance, and helps identify limitations in the learned policies.

By varying these parameters, we create a spectrum of environments that challenge the policies beyond their training distribution, offering a first approximation of the challenges in sim-to-real transfer. This systematic variation allows us to evaluate how well the policies adapt to conditions different from those encountered during training, which is crucial for understanding their potential performance in real-world scenarios where exact environmental conditions may differ from the simulation.

\section{Results \& Discussion} \label{sec:results}
\subsection{Comparing Robustness of First-Order and Zeroth-Order Methods}

Next, we test our intuition that the gradients provided by differentiable simulators in FoPG methods may lead to convergence towards sharp local minima.
These sharp minima are characterized by high sensitivity to small perturbations in both policy and environmental parameters.
We then study the effect of these two critical sensitivities:
\begin{enumerate}
    \item \textbf{Policy Sensitivity:} small perturbations in policy outputs can displace the solution from the optimal point within the sharp local minimum, resulting in significant deterioration of performance.
    \item \textbf{Environmental Sensitivity:} Sharp features are highly sensitive to environmental changes, causing previously optimal solutions to lose effectiveness when conditions deviate from the training scenario.
\end{enumerate}

Theoretical and empirical studies suggest that flatter minima often correspond to more generalizable solutions \cite{hochreiter1997flat}. The inherent noise in zeroth-order policy gradient (ZoPG) methods may naturally bias them towards these more robust solutions, potentially leading to better generalization in diverse scenarios.

To empirically test this hypothesis, we compare the robustness of policies trained using SHAC and PPO in the Ant environment. The Ant environment used in our experiments is a reimplementation of the classical Ant MuJoCo environment \cite{Mujoco} in NVIDIA's DFlex, providing a differentiable simulation platform for our study. 
We examine environmental sensitivity by varying two critical parameters that significantly impact the dynamics of contact-rich scenarios:
contact stiffness and damping which are described in \ref{sub:contac_params}.  

\begin{figure}[tbp]
    \centering
    \begin{subfigure}{0.45\textwidth}
        \centering
        \includegraphics[width=\linewidth]{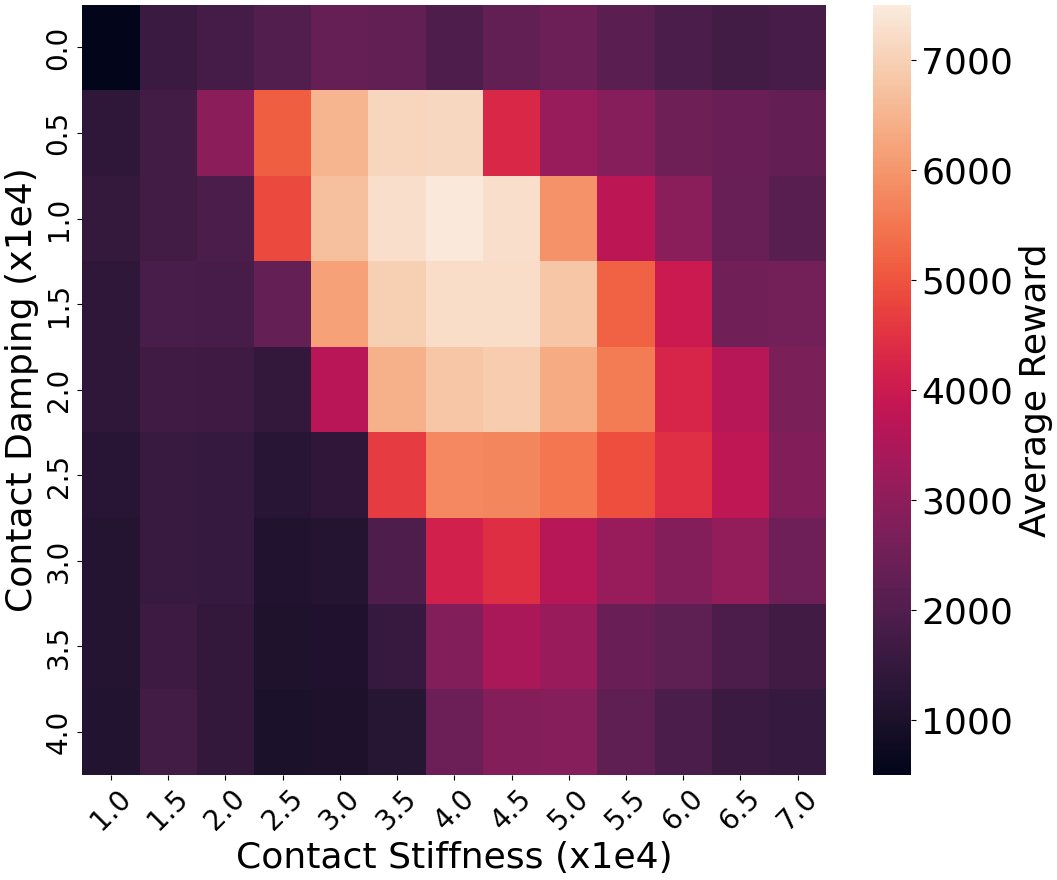}
    \end{subfigure}
    \hfill
    \begin{subfigure}{0.45\textwidth}
        \centering
        \includegraphics[width=\linewidth]{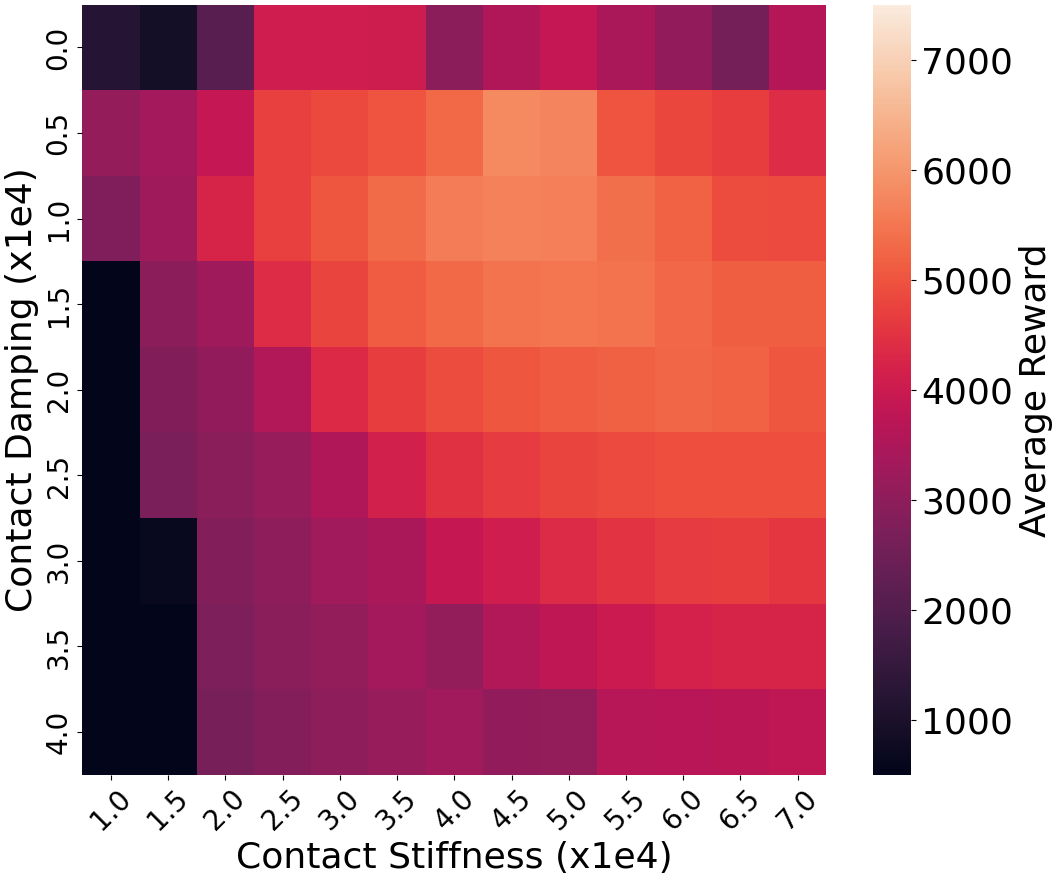}
    \end{subfigure}
    \caption{Average episode reward heatmaps for SHAC (left) and PPO (right) policies under varying contact stiffness ($k_e$) and damping ($k_d$) in the Ant environment.}
    \label{fig:heatmaps}
\end{figure}

The results of our environmental sensitivity experiment, visualized in Figure \ref{fig:heatmaps}, reveal a stark contrast between SHAC and PPO. The heatmaps show the performance of each algorithm across varying levels of contact stiffness and damping in the Ant environment.
While SHAC achieves higher peak rewards under specific parameter combinations as indicated by the brighter regions in the upper heatmap, PPO demonstrates greater robustness to parameter variations.
This is evidenced by the more uniform distribution of rewards across the parameter space in the bottom heatmap. 
However, SHAC policy's performance degrades more rapidly as we move away from its optimal parameter region, indicating a higher sensitivity to changes in these environmental parameters compared to PPO.
In contrast, ZoPG, here exemplified by PPO, relies on stochastic gradient estimates.
Their loss landscape seems to be more consistent across parameters change, indicating to come up with policies with superior generalization capabilities.
The estimation of the policy gradients seems to act as a form of implicit regularization biasing ZoPG methods with a flatter loss landscape.
\subsection{SHAC-ASAM Generalization Capabilities}

\begin{figure}[tbp]
    \centering
    \begin{subfigure}{0.455\textwidth}
        \centering
        \includegraphics[width=\linewidth]{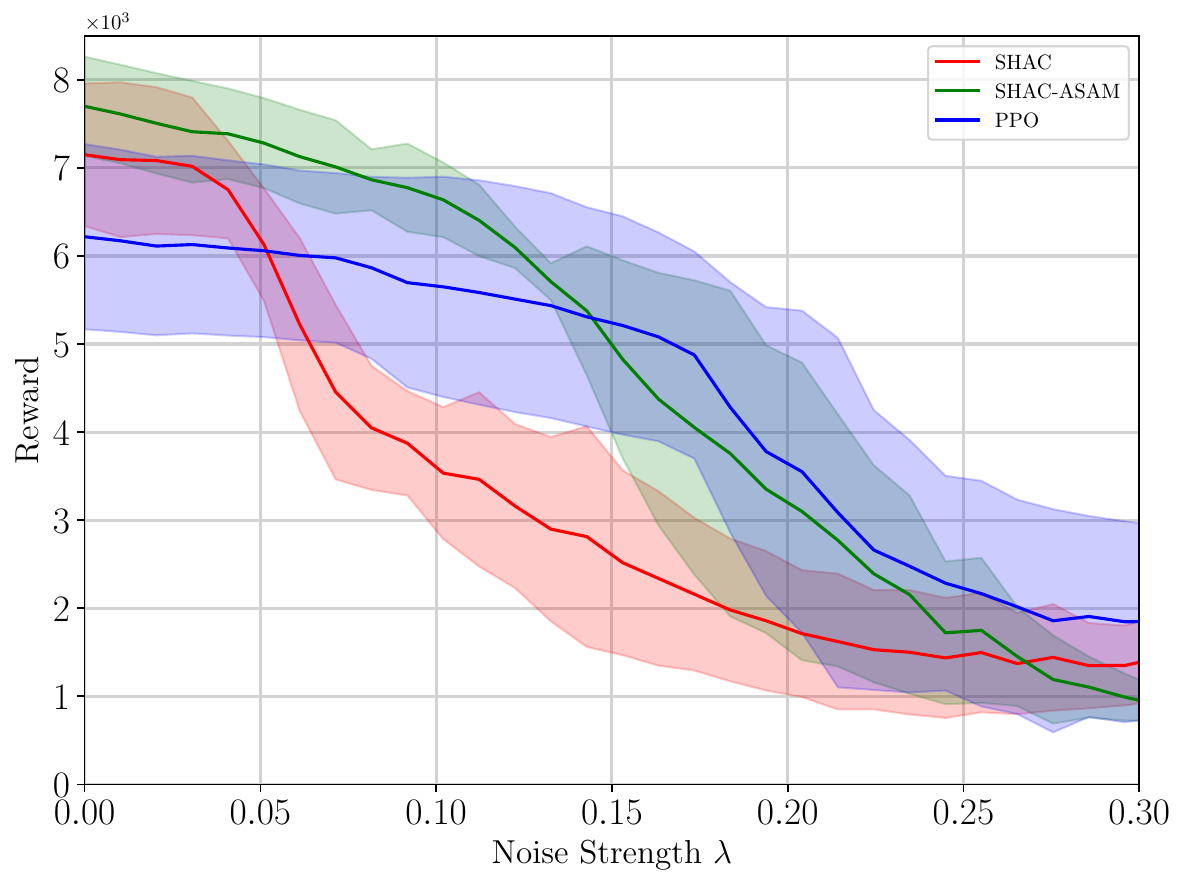}
        \subcaption{Ant}
        \label{fig:action_noise_ant}
    \end{subfigure}
    \hfill
    \begin{subfigure}{0.45\textwidth}
        \centering
        \includegraphics[width=\linewidth]{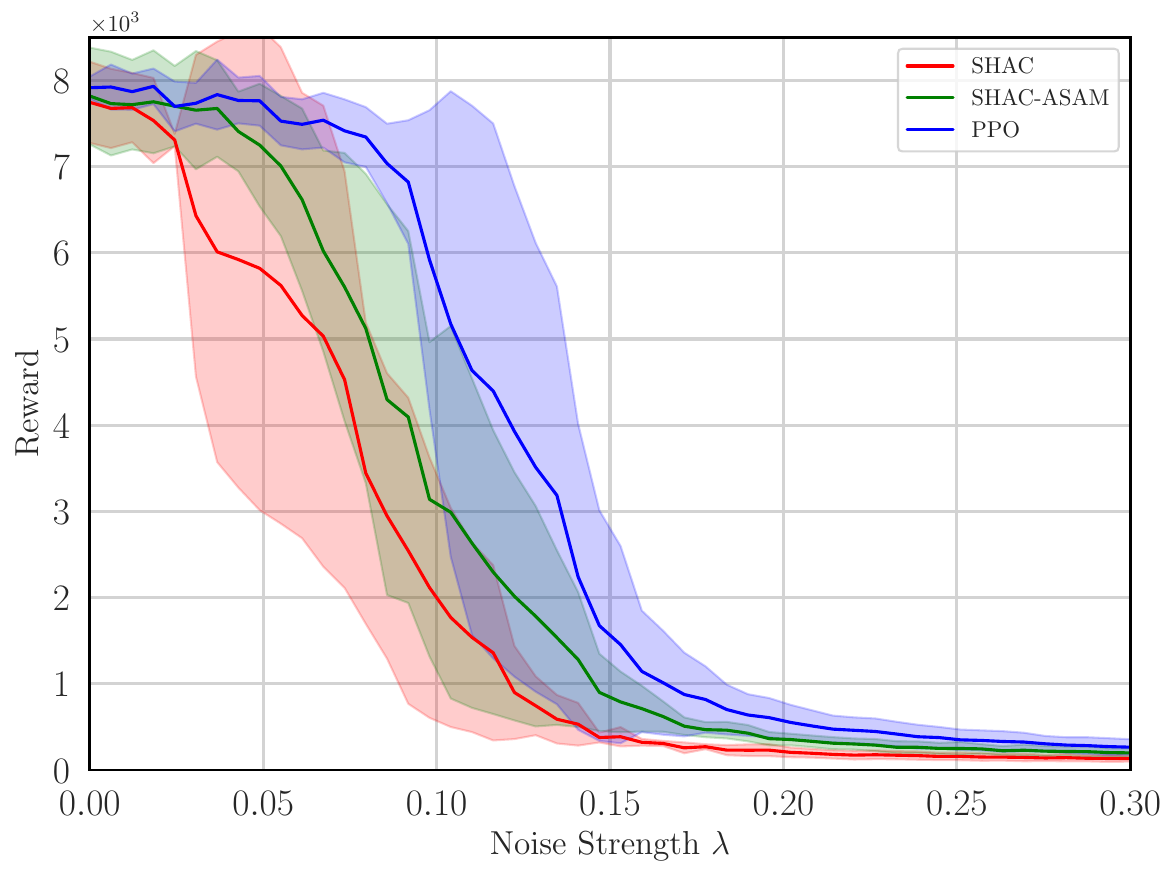}
        \subcaption{Humanoid}
    \end{subfigure}
    \caption{Average episode reward as function of the noise strenght for SHAC, SHAC-ASAM, and PPO. The rewards are averaged over 100 rollouts, from 3 different policies per algorithm. The shades represent the standard deviation of the reward.}
    \label{fig:action_noise}
\end{figure}

\begin{figure}[!htbp]
    \centering
    \begin{subfigure}{0.45\textwidth}
        \centering
        \includegraphics[width=\linewidth]{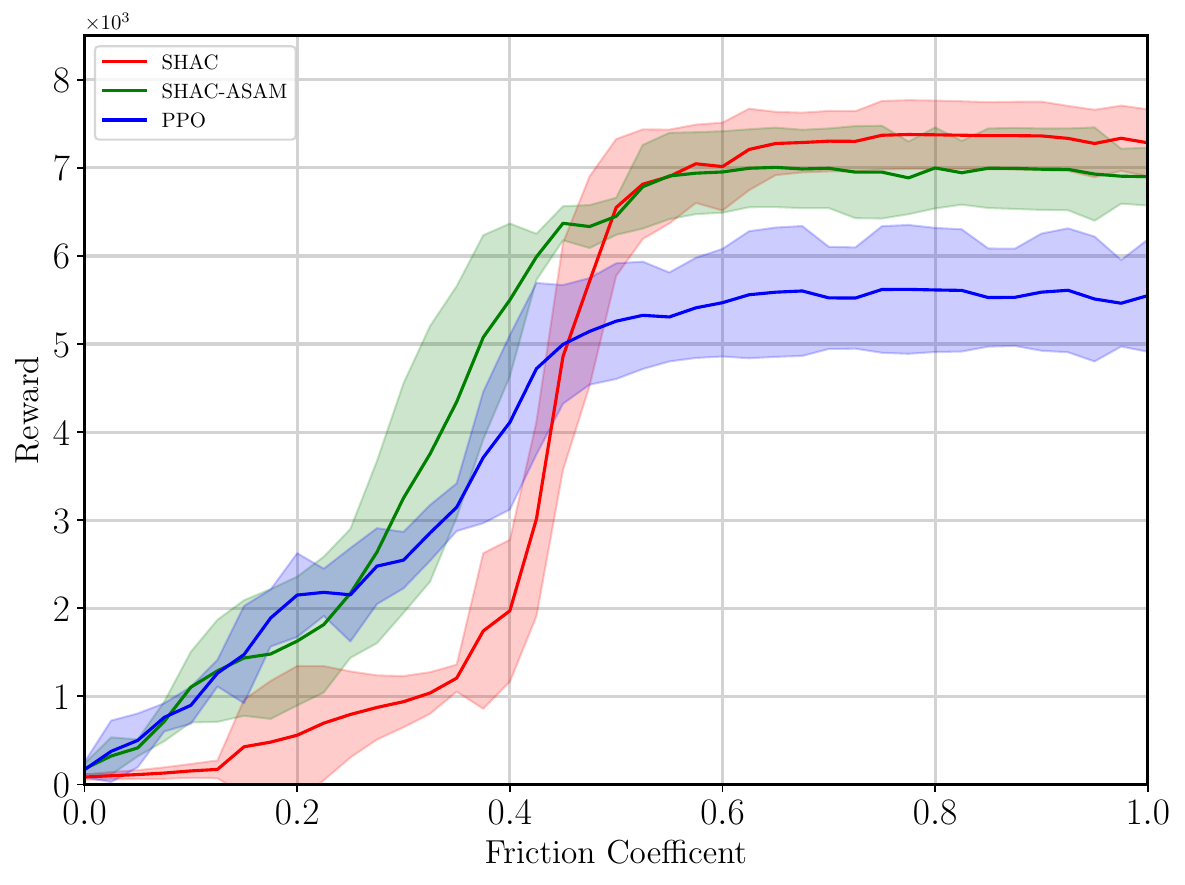}
        \caption{Ant}
        \label{fig:action_noise_ant_2}
    \end{subfigure}
    \hfill
    \begin{subfigure}{0.45\textwidth}
        \centering
        \includegraphics[width=\linewidth]{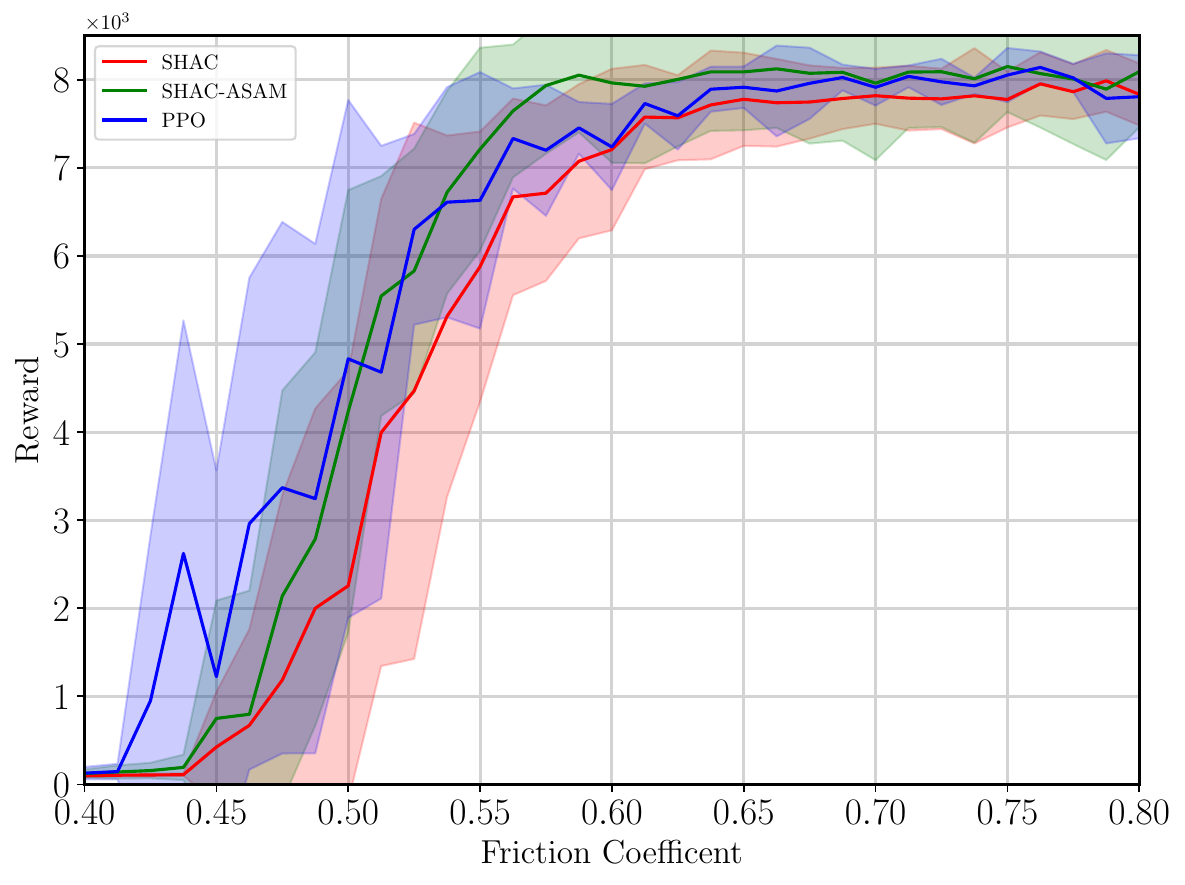}
        \caption{Humanoid}
        \label{fig:action_noise_humanoid}
    \end{subfigure}
    \caption{Average episode reward as a function of the contact Coulomb
    friction for SHAC, SHAC-ASAM, and PPO. The rewards are averaged over 100 rollouts, from 3 different policies per algorithm. The shades represent the standard deviation of the averages of the reward.}
    \label{fig:action_noise_2}
\end{figure}

In this section, we investigate the generalization capabilities of policies trained using SHAC-ASAM by evaluating their performance in terms of rewards and comparing them against the baseline SHAC and PPO algorithms. We assess the policies' robustness under noise perturbations applied to the actuator actions and under varying contact parameters. Additionally, we explore the trade-offs between generalization and performance when choosing the $\rho$ parameter for ASAM, and the trade-off regarding training times of SHAC vs. SHAC-ASAM. The experiments are conducted in the Ant and Humanoid environments.

\subsubsection{Sensitivity to Action Perturbations}

Figure \ref{fig:action_noise_ant} shows the average episode reward versus the level of noise injected into the policy actions as described in Section \ref{sub:noise_actions}. To evaluate the robustness of the policies, we varied the parameter $\lambda$ of the convex combination between the original action and uniform noise. We tested $\lambda$ values ranging from 0 (no noise) to 0.5 (equal weight to original action and noise).

From Figure \ref{fig:action_noise_ant}, it is evident that applying our SHAC-ASAM) algorithm results in a notable improvement in the robustness of SHAC. Compared to the baseline SHAC, SHAC-ASAM maintains a higher reward level than PPO up to a noise strength of 0.15. Moreover, SHAC-ASAM does not experience the steep drop in reward that SHAC exhibits at around 0.05 noise strength.

Similar results can be found for the Humanoid environment as visualized in Fig.~\ref{fig:action_noise_humanoid}. Generally, the Humanoid environment is more unstable and sensitive to perturbations of actions than the Ant environment. However, adding a sharpness-aware optimizer on top of SHAC still leads to a notable improvement in robustness.

\subsubsection{Sensitivity to Contact Parameters Modification}

Next, to evaluate the generalization capabilities of the trained policies under varying environmental conditions, we perturbed the friction coefficient of the environments. As shown in Fig.~\ref{fig:action_noise_ant}, training SHAC with the ASAM optimizer with $\rho = 0.75$ results in better generalization capabilities than training plain SHAC. A similar observation can be made for the humanoid environment in Fig.~\ref{fig:action_noise_humanoid} which is generally less robust to noise than the Ant environment. Still, our sharpness-aware approach can match the generalizability of PPO in this case.

\subsubsection{Balancing Specialization and Generalization in ASAM}

\begin{figure}[h]
    \centering
    \includegraphics[width=0.6\linewidth]{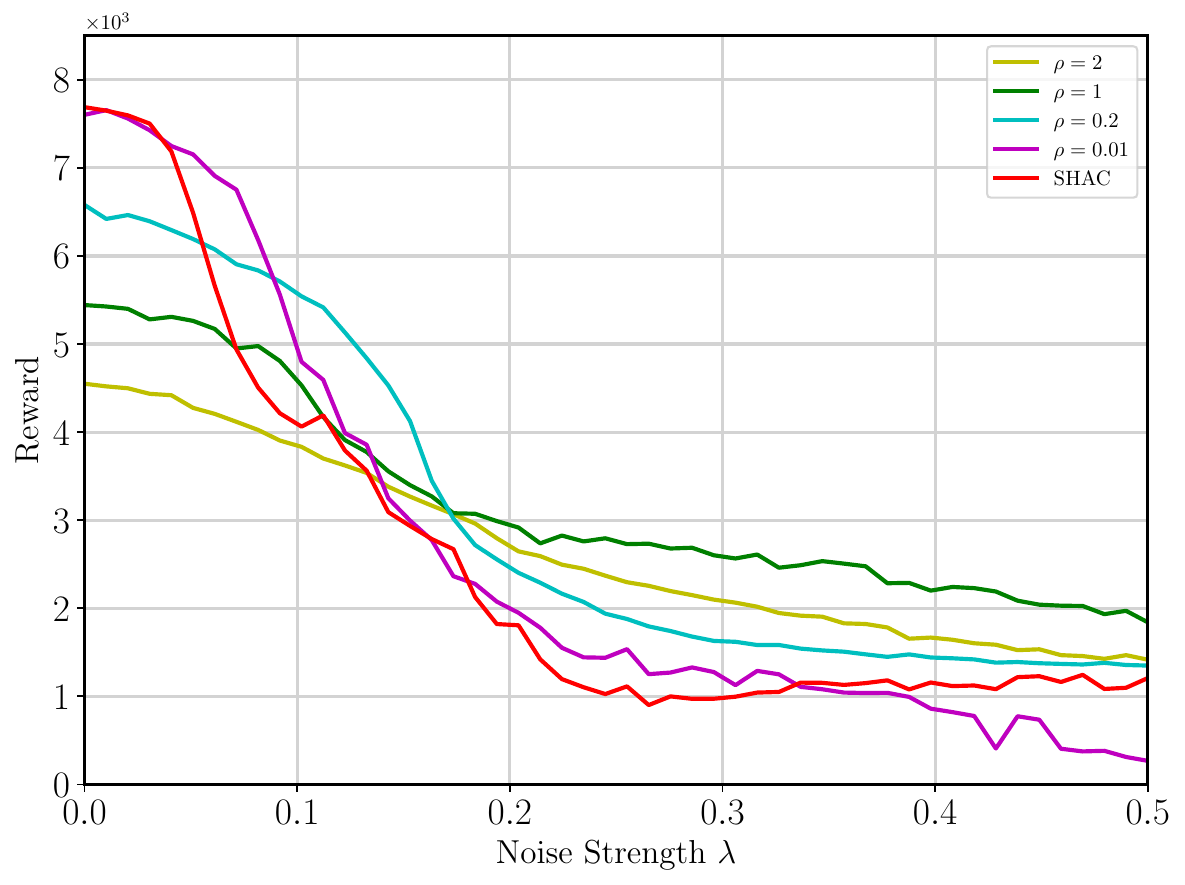}
    \caption{Reward vs Action Noise for policies trained with SHAC-SAM for different $\rho$ values, illustrating the trade-off between performance and generalization}
    \label{fig:rhoplot}
\end{figure}

Fig.~\ref{fig:rhoplot} illustrates the key trade-off in selecting the $\rho$ parameter for SHAC-ASAM, demonstrating how we can tune the balance between generalization and performance within the same sample budget. The results align with our theoretical expectations: larger $\rho$ values lead to flatter minima in the loss landscape, resulting in better generalization but potentially lower peak performance.
For lower $\rho$ values, we observe higher rewards when the testing environment closely matches the training conditions. However, these policies show steeper performance degradation as action noise increases. Conversely, policies trained with larger $\rho$ values exhibit more stable performance across varying noise levels, indicating superior generalization.
We observed that policies with higher $\rho$ values can be trained for more iterations to achieve better overall performance, albeit at the cost of requiring more samples. This suggests that increasing $\rho$ slows down the learning process. Nonetheless, this finding supports our main hypothesis: SHAC-ASAM allows for fine-tuning the generalization-specialization trade-off while maintaining sample efficiency.
These results underscore the flexibility of SHAC-ASAM in adapting to different requirements, whether prioritizing high performance in known conditions or robust generalization in uncertain environments.

\subsubsection{Trade-off Between Training Time and Generalization}
The table given in \ref{tab:experiment_results} depicts the training times for Ant and Humanoid of Vanilla SHAC and our novel method SHAC-ASAM incorporating sharpness-aware optimization. Notably, our approach takes around twice the time of SHAC which aligns with the reported computational complexity in the ASAM paper \cite{kwon2021asam}. However, the higher training cost can potentially be reduced by incorporating more advanced versions of sharpness-aware optimizers that achieve state-of-the-art results in terms of computational cost as mentioned in section \ref{sec:related_work}. 

\begin{table}[H]
\centering
\begin{tabular}{ccc}
\hline
Experiment & SHAC & SHAC-ASAM \\
\hline
Ant & 1436 ± 5 s & 2460 ± 100 s \\
Humanoid & 4400 ± 50 s & 8373 ± 200 s \\
\hline
\end{tabular}
\caption{Training times. Mean ± standard deviation in seconds for SHAC and SHAC-ASAM for Ant and Humanoid. Training done with an NVIDIA GeForce RTX 2080 Ti}
\label{tab:experiment_results}
\end{table}

\section{Conclusion \& Future Work} \label{sec:Conclusion}

In this work, we presented a novel method incorporating sharpness-awareness into differentiable policy optimization. Our work contributes to developing RL algorithms that are both sample-efficient and robust to environmental changes, a crucial step towards successful real-world RL applications.

Simulation experiments on commonly used Mujoco environments demonstrate that our method is effectively improving the robustness of first-order policy optimization methods like SHAC while maintaining most of its sample-efficiency.
In this work we mainly focused on SHAC as a FoPG algorithm, but in principle, our approach of applying ASAM is agnostic of the underlying algorithm and results should transfer to other first-order methods such as PODS or AHAC and other sharpness-aware optimizers.
Hence, we plan to extend and test this approach also for other first-order algorithms in our future work.
While SHAC-ASAM shows an increased generalization
capability, two backward passes are required for each training step.
Another aspect we aim to address in future work is the exploration of less computationally costly sharpness-aware optimizers tailored for applications in robotics.
Finally, we plan to demonstrate the effectiveness of our method on more environments and
sources of perturbation before validating our findings with
sim-to-real testing of our method on real hardware.

{
\section*{Acknowledgements} \label{ack}
    This work stems from a project for the course "Foundations of Reinforcement Learning" offered by the Computer Science Department of ETH Z\"urich.
    We are deeply thankful to {Prof.~Niao~He} and the members of her group for their teaching and our student colleagues for their positive remarks.
}
\bibliographystyle{IEEEtran} 
\bibliography{references} 


\end{document}